\UseRawInputEncoding
\documentclass{article}
\usepackage[a4paper, total={6in, 10in}]{geometry}
\usepackage{graphicx} 
\usepackage{authblk}
\usepackage{ amssymb }
\usepackage{amsmath}
\usepackage{float}
\usepackage{multirow}
\usepackage{adjustbox}
\usepackage{caption}
\usepackage{subcaption}
\usepackage{booktabs}
\usepackage{siunitx}
\usepackage{enumitem}
\usepackage{mathtools,leftindex,tensor,mhchem}
\usepackage{bm}
\usepackage[ruled,vlined]{algorithm2e}

\usepackage{tikz}
\usetikzlibrary{arrows.meta, positioning, calc}
\DeclareSIUnit{\million}{\text{M}}

\usepackage{pifont} 

  \newenvironment{tritemize}
  {\begin{itemize}[label=\textcolor{green!30!gray}{\raisebox{0.2ex}{$\blacktriangleright$}}]}
  {\end{itemize}}

\title{Schr\"odinger-Inspired Time-Evolution for 4D Deformation Forecasting}

\author[1]{Ahsan Raza Siyal}
\author[1]{Markus Haltmeier}
\author[2,3]{Ruth Steiger}
\author[2,3]{Elke Ruth Gizewski}
\author[2,3]{Astrid Ellen Grams}

\affil[1]{Department of Mathematics, University of Innsbruck, Austria}
\affil[2]{Department of Radiology, Medical University
of Innsbruck, Austria}
\affil[3]{Neuroimaging Research Core Facility, Medical
University of Innsbruck, Austria}

\date{}

\begin{document}

\maketitle

\begin{abstract}
Spatiotemporal forecasting of complex three-dimensional phenomena (4D: 3D + time) is fundamental to applications in medical imaging, fluid and material dynamics, and geophysics. In contrast to unconstrained neural forecasting models, we propose a Schr\"odinger-inspired, physics-guided neural architecture that embeds an explicit time-evolution operator within a deep convolutional framework for 4D prediction. From observed volumetric sequences, the model learns voxelwise amplitude, phase, and potential fields that define a complex-valued wavefunction $\psi = A e^{i\phi}$, which is evolved forward in time using a differentiable, unrolled Schr\"odinger time stepper.
This physics-guided formulation yields several key advantages: (i) temporal stability arising from the structured evolution operator, which mitigates drift and error accumulation in long-horizon forecasting; (ii) an interpretable latent representation, where phase encodes transport dynamics, amplitude captures structural intensity, and the learned potential governs spatiotemporal interactions; and (iii) natural compatibility with deformation-based synthesis, which is critical for preserving anatomical fidelity in medical imaging applications. By integrating physical priors directly into the learning process, the proposed approach combines the expressivity of deep networks with the robustness and interpretability of physics-based modeling.
We demonstrate accurate and stable prediction of future 4D states, including volumetric intensities and deformation fields, on synthetic benchmarks that emulate realistic shape deformations and topological changes. To our knowledge, this is the first end-to-end 4D neural forecasting framework to incorporate a Schr\"odinger-type evolution operator, offering a principled pathway toward interpretable, stable, and anatomically consistent spatiotemporal prediction.
\end{abstract}

\section{Introduction}

Modeling and forecasting complex spatiotemporal deformations in 4D (3D + time) is a core challenge in medical imaging, scientific computing, and physical simulation. Accurate motion models underpin registration, motion compensation, radiotherapy planning, longitudinal disease assessment, and quantitative analysis \cite{mcclelland2013respiratory}. In MR guided radiotherapy (MRgRT) and image guided interventions, short horizon prediction on the order of hundreds of milliseconds helps compensate system latency from acquisition, localization, and beam adaptation \cite{kurz2020medical}. Classical approaches grounded in biomechanical, variational, and statistical principles offer realism and physical constraints but are computationally demanding and difficult to scale across anatomies and protocols \cite{mcclelland2013respiratory}. Learning based deformable registration, exemplified by VoxelMorph and related CNN frameworks, dramatically accelerates pairwise alignment \cite{Balakrishnan2019, Dalca2019, Chen2022, siyal, siyal2024siru, siyal2025dare}, yet typically targets instantaneous correspondences rather than explicit multi step forecasting. In parallel, 4D imaging pipelines often exploit cine MRI as a surrogate: dynamic 2D slices are linked to volumetric deformation via population or conditional motion models for whole organ and ROI tracking \cite{mcclelland2013respiratory, wei2021unsupervised}. Numerical and anthropomorphic phantoms such as the 4D XCAT phantom, a realistic computational model of the thorax and abdomen with controllable cardiac and respiratory motion, and post processed deformation vector fields (DVFs), dense voxelwise displacement fields used to warp a reference volume, remain standard for benchmarking under controlled respiratory patterns and for validating invertibility and consistency constraints \cite{segars20104d}.

Early data driven forecasters used temporal recurrence with RNNs and ConvLSTMs \cite{shi2015convolutional}, but autoregressive rollouts accumulate error and struggle under irregular breathing such as sighs, coughs, and apnea \cite{romaguera2023conditional}. Attention based predictors with Transformers mitigate compounding error by forecasting multiple steps in parallel, and achieve better millimeter scale geometric fidelity when combined with deformation based generation \cite{romaguera2023conditional}. Empirically, generating future frames by spatially warping a reference with predicted deformation vector fields (DVFs), which are dense voxelwise displacement fields, yields sharper and more anatomically faithful results than direct intensity regression in medical sequences \cite{romaguera2023conditional}. More broadly, deformation forecasting is closely related to optical and scene flow via the transport equation
$ \partial_t I(\mathbf{r}, t) + \mathbf{v}(\mathbf{r}, t)\cdot\nabla_{\mathbf{r}} I(\mathbf{r}, t) = 0
$,
where $I(\mathbf{r}, t)$ is the image intensity at spatial location $\mathbf{r}\in\mathbb{R}^3$ and time $t$, $\mathbf{v}(\mathbf{r}, t)$ is the Eulerian velocity field, and $\nabla_{\mathbf{r}}$ is the spatial gradient; the dot denotes the Euclidean inner product. This motivates estimating either the velocity field $\mathbf{v}$ or an associated deformation map $\boldsymbol{\phi}_{t\to t+\Delta t}$ that transports intensities forward in time.
Despite strong empirical performance, purely data driven models offer limited interpretability and can be unstable for long horizons. Physics guided and motion structure aware designs alleviate this by encoding deformation as DVFs, encouraging diffeomorphic behavior such as stationary velocity fields in the log domain, LDDMM, and SyN, and constraining temporal evolution with inductive biases \cite{Beg2005, Avants2008, Rueckert1999}. Complementary efforts in physics informed and operator learning models incorporate PDE structure and stable integration into neural architectures \cite{wang2024pinn,kovachki2023neural}. Disentangling temporal dynamics from spatial warping further preserves high frequency anatomy and supports downstream tracking.

The time dependent Schr\"odinger equation \cite{schrodinger1926, Griffiths2018} governs the unitary evolution of a complex wavefunction $\psi(\mathbf{r}, t)$ depending on spatial coordinate $\mathbf{r}$ and time $t$:
\begin{equation}
i\hbar\,\partial_t \psi(\mathbf{r}, t)
= \left(-\frac{\hbar^2}{2m}\nabla^2 + V(\mathbf{r}, t)\right)\psi(\mathbf{r}, t),
\label{eq:schrodinger}
\end{equation}
where $\psi(\mathbf{r}, t)$ is the complex wavefunction, $\mathbf{r}$ is the spatial coordinate, $t$ is time, $m$ is the particle mass, $V(\mathbf{r}, t)$ is the potential, $\nabla^2$ is the spatial Laplacian, and $\hbar$ is the reduced Planck constant.
 Writing $\psi = A\,e^{i\phi}$ (amplitude--phase form) connects to transport: the amplitude $A$ relates to a density proxy $A^2$, while the phase $\phi$ induces a velocity prior via its gradient, $\mathbf{v}\propto\nabla\phi$ (Madelung transform), providing a compact and interpretable representation of dynamics. We adopt this structure as a modeling prior rather than a strict physical law: our numerical scheme does not enforce exact probability or energy conservation, but the induced constraints promote smoother, more stable temporal evolution than unconstrained predictors.

To address these challenges, we propose a Schr\"odinger inspired 4D deformation forecaster that encodes observed volumes into voxelwise amplitude (structural morphology), voxelwise phase (transport), and voxelwise potential (external influences). These components form a complex valued wavefunction $\psi = A\,e^{i\phi}$, which we evolve forward in time via a differentiable unrolled Schr\"odinger stepper. For efficiency and end to end training, we use an explicit predictor corrector approximation to the Crank Nicolson scheme. While not strictly unitary, this update constrains the dynamics and reduces drift over long horizons. Our approach offers at least three benefits:
\begin{tritemize}
    \item \textsc{Interpretable latent structure:} Disentangling amplitude, phase, and potential yields a physically motivated representation of motion in place of black box latent states.
    \item \textsc{Physics inspired stability:} The complex representation and time evolution operator act as natural regularizers, improving long horizon stability compared to purely autoregressive updates.
    \item \textsc{Compatibility with deformation based synthesis:} Because we evolve a latent state and then decode DVFs or synthesize intensities by warping, the method aligns with DVF based frame generation known to preserve anatomy.
\end{tritemize}
Unlike prior forecasters based primarily on recurrent or Transformer architectures, our approach embeds a Schr\"odinger type evolution operator with an interpretable amplitude, phase, and potential factorization, bridging expressive neural networks with physics guided constraints for stable, long horizon 4D deformation forecasting. We validate on synthetic 4D benchmarks that emulate realistic shape deformations and topological changes, and we analyze boundary conditions, discretization, and ablations relevant to stability and accuracy.

\section{Related Work}

Deformation modeling and temporal motion prediction have evolved from classical biomechanical and variational formulations to fast learning based pipelines that support 4D applications in medical imaging and beyond. Early respiratory motion models framed the problem statistically and biomechanically, emphasizing inter and intra subject variability for tracking and dose delivery, and established evaluation practices that remain influential \cite{mcclelland2013respiratory}. Learning based deformable registration, exemplified by VoxelMorph, dramatically reduced computation while maintaining accuracy, enabling near real time pairwise alignment but without explicit temporal forecasting \cite{Balakrishnan2019, wei2021unsupervised}. In MR guided radiotherapy, system latencies on the order of hundreds of milliseconds make ahead of time motion prediction essential; the clinical context and challenges of MRgRT motivate surrogate driven and predictive solutions \cite{kurz2020medical, henke2018phase}. Population and conditional motion models relate partial observations such as 2D cine slices to high dimensional deformation vector fields (DVFs) for both global whole field and local ROI estimation, with cine imaging acting as a robust surrogate for volumetric motion \cite{mcclelland2013respiratory, romaguera2023conditional}. Benchmarks increasingly rely on numerical and anthropomorphic phantoms such as the 4D XCAT and on post processed DVFs that ensure invertibility and consistency, supporting controlled experiments across breathing patterns and acquisition protocols \cite{segars20104d, eiben2020consistent}.

On the temporal modeling side, recurrent architectures such as RNNs and ConvLSTMs were among the first to exploit sequence structure for medical video and motion traces, but their auto regressive nature can accumulate error over longer horizons and under irregular breathing \cite{shi2015convolutional, romaguera2021predictive}. Attention based predictors using Transformers enable parallel multi step forecasting with learnable queries, improving robustness and millimeter scale geometric accuracy when integrated with 4D motion models for volumetric DVF prediction and target tracking \cite{romaguera2023conditional, wang2021transformer, cai2021deep}. In parallel, registration aware tracking and Siamese paradigms have been adapted to 3D and to medical modalities, sometimes augmented with affine priors or motion fields, though deformation centric formulations remain advantageous in motion rich anatomies \cite{nam2016mdnet, miao2016cnn, oktay2018attention}. Acquisition strategies such as navigator echoes, self sorting, and compressed sensing enable long cine sessions and large 4D datasets for training, while deployment uses only real time 2D cine plus a static volume, aligning with practical MRgRT workflows \cite{vonsiebenthal2007mr, romaguera2023conditional}. Recent studies show that spatial transform based future frame generation, warping a reference with predicted DVFs, yields sharper and more anatomically faithful predictions than direct pixel regression, and that combining global motion models with local, latent gated refinement reduces tracking error and mitigates drift \cite{romaguera2023conditional, wei2021unsupervised}. Taken together, this literature motivates deformation first, forecast capable models that are compatible with clinical latencies, leverage cine surrogates, and maintain stability over multi step horizons.

\section{Methodology}

\subsection{Problem Setup}

Let $X_{t-k+1:t} \triangleq (X_{t-k+1}, \dots, X_{t})$ denote an ordered history of $k$ volumetric observations with $X_{s} \in \mathbb{R}^{H \times W \times D}$, where each $X_{s}$ is a 3D scan. We consider 4D forecastin, which consists of learning a forecasting function
\begin{equation} \label{eq:forecasting}
F_\theta \colon X_{t-k+1:t} \mapsto \hat{X}_{t+1},
\end{equation}
parameterized by $\theta$, that predicts the next volume $\hat{X}_{t+1}$ given the observed history.
During supervised training, the true future volume $X_{t+1}$ is available and the model is optimized to minimize a prediction risk
\begin{equation} \label{eq:forecast_loss}
\mathcal{L}(\theta) \;=\; \ell\!\left(\hat{X}_{t+1},\, X_{t+1}\right),
\end{equation}
where $\ell$ penalizes deviations of predictions from ground truth; common choices include mean squared error (MSE), SSIM, or a task specific metric. This problem is particularly challenging because volume deformations are nonlinear, high dimensional, and often governed by complex physical processes such as tumor growth, respiratory motion, or material dynamics. Conventional approaches treat 4D forecasting as a purely data-driven regression problem or rely on recurrent or attention-based networks, which may accumulate drift and lack interpretability in long-horizon prediction.

We propose to learn a physics-inspired forecaster that maps past volumes $X_{t-k+1:t}$ to a forecast $\hat{X}_{t+1}$ by minimizing a supervised regularized prediction loss
\begin{equation}
\min_{\theta}\;
\mathbb{E}_{(X_{t-k+1:t},\, X_{t+1}) \sim \mathcal{D}}\Big[
\|F_\theta(X_{t-k+1:t}) - X_{t+1}\|_2^2 + \lambda\,\mathcal{R}\big(F_\theta(X_{t-k+1:t})\big)
\Big],
\end{equation}
where the expectation is over training samples from $\mathcal{D}$. The mapping $F_\theta$ is constrained by a Schr{\"o}dinger-inspired latent structure where $\hat{X}_{t+1} = |\psi^{(N)}|^2$ and $\psi(\mathbf{r}, t) = A(\mathbf{r}, t)\exp(i\Phi(\mathbf{r}, t))$ with amplitude $A$, phase $\Phi$, and potential $V$ predicted by convolutional encoder--decoder heads. The complex wavefunction then evolves through $N$ steps of an unrolled approximate Crank--Nicolson scheme under the Hamiltonian $H\psi = -\nabla^2\psi/2 + V\psi$.

\subsection{Schr{\"o}dinger-Inspired Time Evolution}

We begin from the time-dependent Schr{\"o}dinger equation \cite{schrodinger1926,Griffiths2018} governing the temporal evolution of quantum-mechanical systems. As a simplification of the general form in \eqref{eq:schrodinger}, we adopt normalized units $\hbar = 1$, $m = 1$, yielding
\begin{equation}\label{eq:schrodinger-norm}
i\,\frac{\partial \psi(\mathbf{r}, t)}{\partial t}
= \left(-\tfrac{1}{2}\nabla^2 + V(\mathbf{r}, t)\right)\psi(\mathbf{r}, t).
\end{equation}
This simplified formulation retains the essential structure: the Laplacian encodes the kinetic term, while $V(\mathbf{r}, t)$ modulates the evolution through external or internal driving forces. In quantum mechanics, $|\psi|^2$ represents probability density and the phase $\arg(\psi)$ encodes temporal and spatial oscillations. Inspired by this, we adopt a complex representation and evolution for deformation forecasting. This allows us to disentangle morphology (through amplitude), dynamics (through phase), and driving forces (through potential), while providing a natural way to embed oscillatory and structured temporal behavior. This representation acts as a compact, interpretable latent state for spatiotemporal deformation.

Given an input sequence of $k$ volumetric frames $X_{t-k+1:t}$, we concatenate them into a tensor $\mathbf{X}_t \in \mathbb{R}^{k \times H \times W \times D}$ and process it with a 3D convolutional encoder $E_\theta(\cdot)$. The encoder is a three-stage 3D convolutional backbone with increasing feature capacity (16, 32, 64 channels). Each stage applies two 3D convolutions with batch normalization and ReLU activation, enabling hierarchical extraction of spatiotemporal deformation features from the volumetric input:
\begin{equation}
e_\theta(t) \;=\; E_\theta(\mathbf{X}_t) \;\in\; \mathbb{R}^{C \times H \times W \times D},
\end{equation}
where $C$ denotes the number of latent channels. From this shared representation, three convolutional heads $f_A$, $f_\Phi$, and $f_V$ extract the amplitude, phase, and potential fields, respectively:
\begin{align*}
A(\mathbf{r}, t) &= \mathrm{ReLU}\!\big(f_A(e_\theta(t))(\mathbf{r})\big), \\
\Phi(\mathbf{r}, t) &= \pi \cdot \tanh\!\big(f_\Phi(e_\theta(t))(\mathbf{r})\big), \\
V(\mathbf{r}, t) &= \tanh\!\big(f_V(e_\theta(t))(\mathbf{r})\big).
\end{align*}
Here, the amplitude field is nonnegative and represents structural intensity; the phase field is bounded within $[-\pi, \pi]$ and encodes temporal and oscillatory deformation dynamics; and the potential field, bounded for numerical stability, modulates the Hamiltonian evolution. The three fields are combined into a complex-valued latent wavefunction and Hamiltonian operator:
\begin{align} \label{eq:psi-encoded}
\psi(\mathbf{r}, t) &= A(\mathbf{r}, t)\,\exp\!\big(i\,\Phi(\mathbf{r}, t)\big), \\
\label{eq:H-encoded}
H\psi(\mathbf{r}, t) &= -\tfrac{1}{2}\,\nabla^2 \psi(\mathbf{r}, t) + V(\mathbf{r}, t)\,\psi(\mathbf{r}, t).
\end{align}
The basic idea is to evolve $\psi$ defined by \eqref{eq:psi-encoded} under the Schr{\"o}dinger equation \eqref{eq:schrodinger-norm} with Hamiltonian \eqref{eq:H-encoded} to produce the forecast $\hat{X}_{t+1}$ via approximate Schr{\"o}dinger evolution defined below.

The decomposition into amplitude $A$, phase $\Phi$, and potential $V$ provides a structured, interpretable latent representation for deformation forecasting. The amplitude field $A \ge 0$ captures morphological intensity patterns, preserving anatomical fidelity in volumetric data. The phase field $\Phi \in [-\pi,\pi]$ encodes temporal and directional dynamics, enabling a bounded and stable representation of deformation trajectories. The potential field $V$ acts as a modulation term, steering local evolution and embedding deformation-driving forces into the dynamics. Together, these fields define a complex wavefunction \eqref{eq:psi-encoded} that evolves with greater stability and interpretability than purely black-box features, making the approach well-suited for long-horizon spatiotemporal prediction.

\subsection{Approximate Schr{\"o}dinger Evolution}

To evolve $\psi(\mathbf{r}, t)$ in time, we introduce a differentiable evolution operator inspired by the Crank--Nicolson discretization of the Schr{\"o}dinger equation:
\begin{equation}
\left(I + \frac{i\,\Delta t}{2}\,H\right)\psi^{(n+1)}
= \left(I - \frac{i\,\Delta t}{2}\,H\right)\psi^{(n)} \,,
\end{equation}
where $H = -\tfrac{1}{2}\nabla^2 + V$ is the discrete Hamiltonian operator. This scheme is unitary (for real $V$ and a Hermitian discretization of $H$) and preserves the $L^2$ norm of $\psi$, but solving the implicit system is computationally expensive in high dimensions.

Instead, our implementation uses an explicit predictor–corrector approximation \cite{smith1985,ames1992} for the ODE $\partial_t \psi = -i\,H\psi$:
\begin{equation}
\psi^{(n+1/2)} \;=\; \psi^{(n)} \;-\; \frac{i\,\Delta t}{2}\, H\,\psi^{(n)},
\end{equation}
\begin{equation}
\psi^{(n+1)} \;=\; \psi^{(n)} \;-\; i\,\Delta t\, H\,\psi^{(n+1/2)}
\end{equation}
with initialization $\psi^{(0)}(\mathbf{r})  = \psi(\mathbf{r}, t)$.
Here, the Hamiltonian is applied using a discrete 3D Laplacian with periodic boundary conditions, implemented efficiently with tensor rolling operations. Although this explicit scheme introduces small norm drift, it retains the structural benefits of Schr{\"o}dinger-inspired dynamics and is differentiable end-to-end for training. The evolution is unrolled for $N$ steps (with $\Delta t = 1/N$) to simulate temporal progression, yielding the forecast $\psi(\mathbf{r}, t+1)$.

\subsection{Forecasting and Reconstruction}

After $N$ unrolled steps, the evolved wavefunction $\psi^{(N)}$ is used to reconstruct the predicted volume as its normalized intensity:
\begin{equation}
\hat{X}_{t+1}(\mathbf{r}) \;=\; \frac{\big|\psi^{(N)}(\mathbf{r})\big|^2}{\max_{\mathbf{r}} \big|\psi^{(N)}(\mathbf{r})\big|^2 + \epsilon},
\end{equation}
with a small $\epsilon > 0$ for numerical stability. The model is trained to minimize the discrepancy between the prediction and the ground-truth volume using a composite loss:
\begin{equation}
\mathcal{L} \;=\; \big\| \hat{X}_{t+1} - X_{t+1} \big\|_2^2 \;+\; \lambda\,\mathcal{R}(\hat{X}_{t+1}),
\end{equation}
where the first term is the mean squared error (MSE) reconstruction loss, and $\mathcal{R}(\cdot)$ denotes a regularization term. In this work, we use a total variation (TV) penalty, which encourages smoothness in the predicted deformation field and suppresses spurious high-frequency artifacts. TV regularization is particularly important in medical and scientific imaging applications, as it helps maintain anatomical fidelity and reduces noise amplification during long-horizon forecasting.

This methodology integrates physics-inspired priors into deep learning for deformation forecasting, offering three key benefits:
(i) interpretability through disentangled amplitude, phase, and potential fields,
(ii) improved temporal stability via Schr{\"o}dinger-inspired evolution with an approximate Crank--Nicolson update, and
(iii) compatibility with deformation-based forecasting pipelines used in 4D medical imaging and scientific simulations.
The result is a framework that balances expressive power, interpretability, and stability, providing accurate long-horizon volumetric predictions.

\section{Experiments and Results}

We present a comprehensive evaluation of the proposed Schr\"odinger-inspired 4D deformation forecasting framework on synthetic volumetric datasets, which provide ground-truth dynamics for rigorous testing. Unlike clinical data, synthetic sequences allow controlled probing of stability, robustness, and long-horizon performance. We assess voxel-level reconstruction fidelity (MSE, PSNR, SSIM), segmentation accuracy under varying thresholds (Dice), and boundary precision via surface distances, while spectral error decomposition separates global drift from fine-detail preservation. Interpretability is examined through visualization of learned potentials, amplitudes, and energy densities, and computational trade-offs are benchmarked across unroll depths. This multifaceted evaluation establishes the framework’s accuracy, robustness, efficiency, and interpretability, providing a solid basis for future medical and scientific applications.

\subsection{Synthetic Dataset Generation}

To evaluate our framework under controlled conditions, we generated synthetic 4D and 3D volumetric datasets designed to probe different aspects of forecasting. The 4D dataset mimics tumor-like growth with temporal evolution, irregular boundaries, and stochastic variations, providing a realistic setting to study long-horizon deformation dynamics. In contrast, the 3D dataset is used to analyze the network’s capability to capture rotations, translations, and diverse shape deformations, thereby testing the generalizability of the approach. All datasets consist of sequences of length $6$ ($t=0 \dots 5$), where the first $5$ frames are used as input and the $6$-th frame serves as the supervised prediction target. The underlying dynamics combine growth rate, rotational drift, irregular perturbations, and random shifts of the object center. For the 4D experiments, each sequence had dimension $5 \times 64 \times 64 \times 64$, with $1200$ volumes used for training and $400$ for testing. For the 3D experiments, each sequence had dimension $5 \times 128 \times 128$, with $2500$ training volumes and $1000$ test samples. This design yields realistic yet fully controlled spatiotemporal patterns, enabling systematic evaluation of accuracy, stability, and robustness across different deformation scenarios.

\begin{minipage}{0.48\linewidth}
\begin{algorithm}[H]
\caption{Synthetic 3D-Dataset}
\KwIn{Number of samples $N$, volume size $H \times W \times D$, timesteps $T=6$}
\For{$i=1$ to $N$}{
 Sample dynamics: growth\_rate, rotation\_speed, irregularity, center\_shift\;
 \For{$t=0$ to $T-1$}{
   Compute radius: $r = r_0 + t \cdot \text{growth\_rate}$\;
   Compute spherical coordinates $\theta,\phi$ around shifted center\;
   Apply perturbation: $r_{\text{eff}} = r(1 + \text{irregularity}\cdot \sin(3(\theta + \text{rotation\_speed} \cdot t)) \cos(2\phi))$\;
   Construct volume: $\exp(-((\text{dist}-r_{\text{eff}})^2)/5)$\;
   Smooth with Gaussian filter\;
 }
 Return $(X_{0:4}, X_5)$
}
\end{algorithm}
\end{minipage}
\hfill
\begin{minipage}{0.48\linewidth}
\begin{algorithm}[H]
\caption{Synthetic 2D-Dataset}
\KwIn{Number of samples $N$, image size $H \times W$, timesteps $T=6$}
\For{$i=1$ to $N$}{
 Sample dynamics: growth\_rate, rotation\_speed, irregularity, center\_shift\;
 \For{$t=0$ to $T-1$}{
   Compute radius: $r = r_0 + t \cdot \text{growth\_rate}$\;
   Compute angle field $\theta = \arctan2(Y-c_y, X-c_x)$\;
   Apply perturbation: $r_{\text{eff}} = r(1 + \text{irregularity}\cdot \sin(3(\theta + \text{rotation\_speed} \cdot t)))$\;
   Construct mask: $\exp(-((\text{dist}-r_{\text{eff}})^2)/10)$\;
   Smooth with Gaussian filter, normalize to $[0,1]$\;
 }
 Return $(X_{0:4}, X_5)$
}
\end{algorithm}
\end{minipage}

\begin{figure}[h!]
    \centering
    \includegraphics[width=0.9\textwidth]{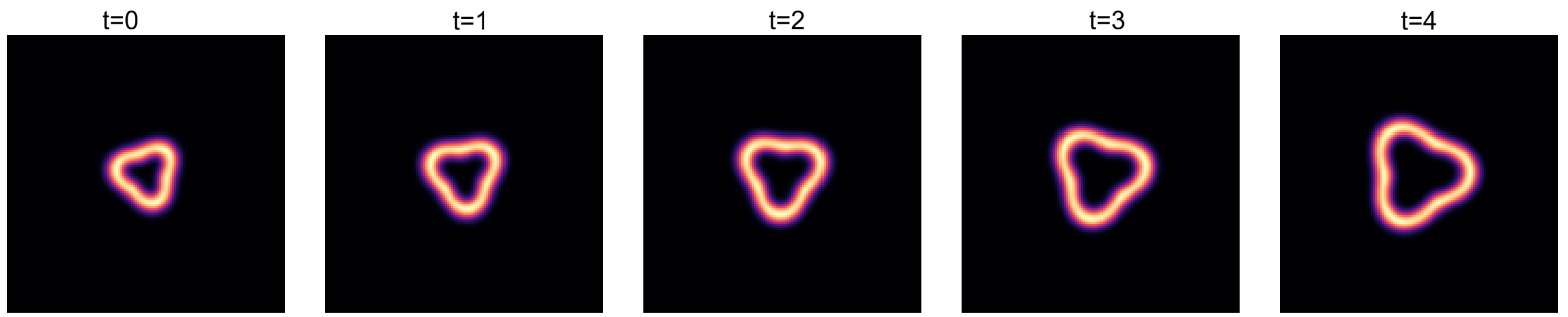}
    \caption{Examples of 2D dataset samples at different timepoints, illustrating volumetric irregularity and realistic deformation dynamics.}
    \label{fig:2d_dataset}
\end{figure}

\begin{figure}[h!]
    \centering
    \includegraphics[width=1\textwidth]{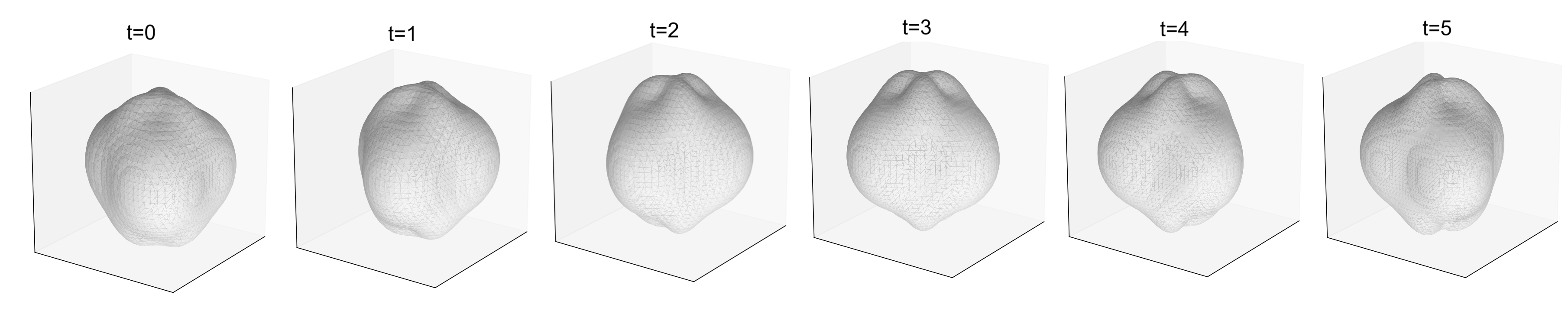}
    \caption{Isosurface visualization of random 3D dataset sample at different timepoint, showing diverse tumor-like morphologies with irregular boundaries.}
    \label{fig:3d_isosurface}
\end{figure}

To better illustrate the synthetic datasets used in this study, we provide representative visualizations of both 2D and 3D cases. 
Figure~\ref{fig:2d_dataset} shows examples of 2D shapes across different timepoints, highlighting the irregular boundaries and rotational variations that challenge forecasting models. 
For the 3D dataset, Figure~\ref{fig:3d_isosurface} depicts isosurface renderings of random sample at different timepoints, which capture realistic volumetric growth and morphological irregularities. 
These figures confirm that the synthetic datasets successfully mimic realistic spatiotemporal deformation patterns, enabling systematic evaluation of forecasting methods.

\subsection{Effect of Unroll Depth on Forecasting Accuracy and Stability}
Table~\ref{tab:metrics_by_unroll} reports the effect of varying the number of unrolled Schr\"odinger evolution steps on forecast accuracy. 
Moving from 10 to 50 unrolls improves all metrics substantially: SSIM increases by $+0.0086$ (a relative gain of $0.87\%$), PSNR rises by $+9.2$~dB, MSE decreases by a factor of $8.5$, and Dice improves by $+0.036$. 
These gains demonstrate that deeper unrolling captures more of the deformation dynamics while remaining stable. 
However, pushing the horizon to 100 steps leads to severe degradation: SSIM drops by $-0.030$, PSNR decreases by $-18.1$~dB, MSE increases by $61.7\times$, and Dice falls by $-0.397$. 
This behavior is consistent with the accumulation of numerical approximation error and non-unitary drift, confirming that moderate unroll depths strike the best balance between expressivity and stability. 
In particular, $50$ unrolls emerged as the optimal trade-off for long-horizon volumetric forecasting.

\begin{table}[thb!]
\centering
\caption{Effect of unrolled Schr\"odinger evolution steps on forecast accuracy (mean over $N$ cases). Best values in \textbf{bold}.}
\label{tab:metrics_by_unroll}
\begin{tabular}{rccccc}
\toprule
Unroll & SSIM $\uparrow$ & PSNR (dB) $\uparrow$ & MSE $\downarrow$ & Dice $\uparrow$  \\
\midrule
10  & 0.9888 & 34.87 & 1.92e-04 & 0.9511  \\
20  & 0.9940 & 37.04 & 1.48e-04 & 0.9548  \\
50  & \textbf{0.9973} & \textbf{44.09} & \textbf{2.26e-05} & \textbf{0.9866} \\
100 & 0.9673 & 26.01 & 1.39e-03 & 0.5894 \\
\bottomrule
\end{tabular}
\end{table}

\begin{figure}[thb!]
\centering
\includegraphics[width=0.6\linewidth]{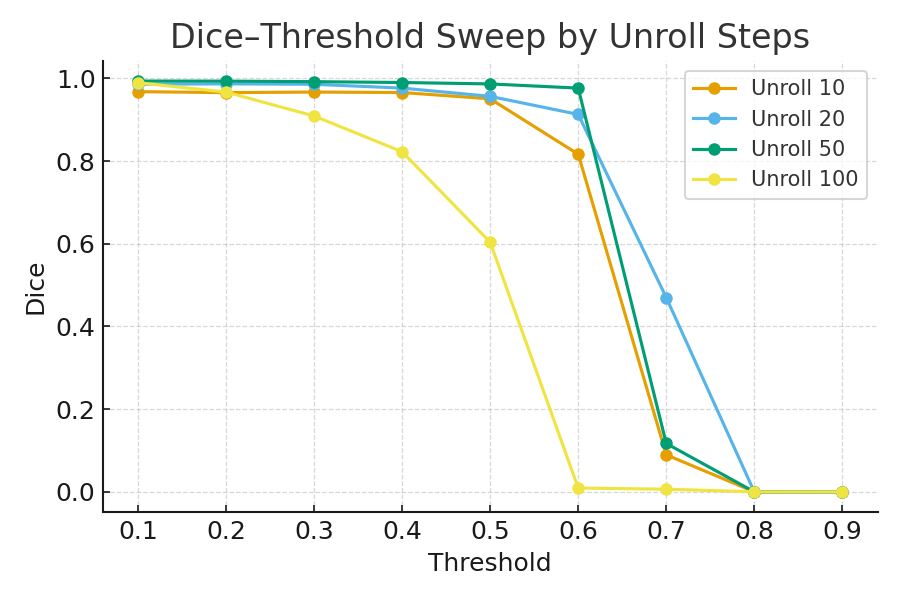}
\caption{Dice--threshold sweep for different unroll steps. Curves illustrate operating characteristics and robustness to the choice of binarization threshold $\tau$.}
\label{fig:dice_sweep}
\end{figure}

In addition to reconstruction fidelity, we examined how segmentation performance varied with the binarization threshold $\tau$ applied to the predicted intensity maps. 
Table~\ref{tab:metrics_by_unroll} reports Dice scores at a fixed threshold of $\tau=0.5$, which serves as a consistent operating point across unroll settings and allows direct comparison of forecasting accuracy. 
However, the choice of threshold can substantially influence segmentation outcomes, motivating a broader analysis. 
Figure~\ref{fig:dice_sweep} shows Dice scores across a range of thresholds for different unroll depths. 
The curves reveal that $50$ unrolls not only achieve the highest peak Dice ($0.9938$) but also maintain performance within $99\%$ of this maximum over a wide interval ($\tau \in [0.10, 0.50]$). 
This plateau demonstrates strong robustness to threshold selection, which is valuable in practice where calibration may be imperfect. 
By contrast, shorter unrolls (e.g., $10$ steps) exhibit narrower ranges of robustness, while excessively long horizons ($100$ steps) suffer reduced overall Dice despite high recall at low thresholds. 
These results confirm that the Schr\"odinger-inspired forecasting model achieves stable, high-fidelity predictions at moderate unroll depths, both in continuous regression metrics and in discrete segmentation quality.

\begin{figure}[t]
\centering
\includegraphics[width=0.9\linewidth]{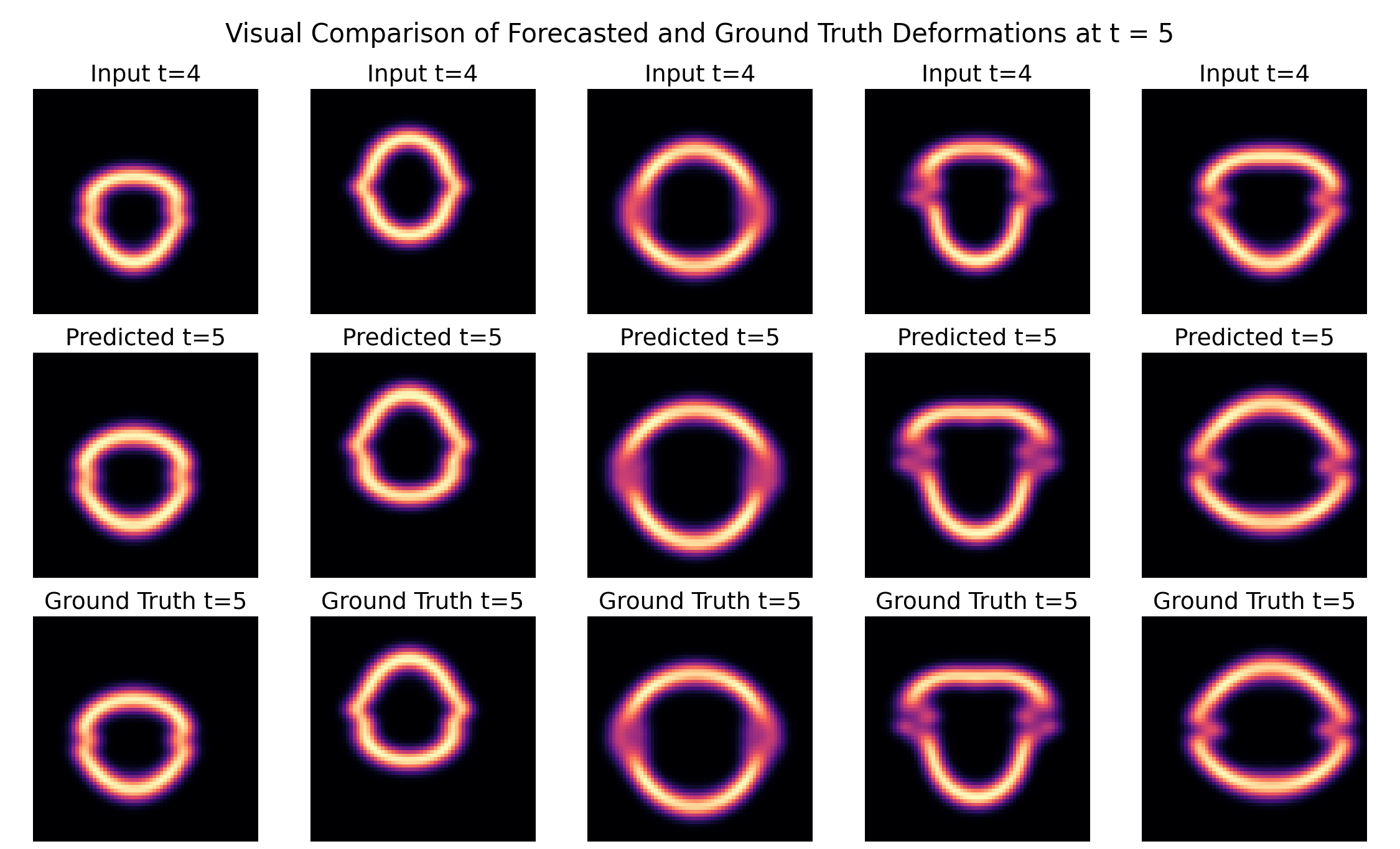}
\caption{Visual comparison of forecasted and ground truth deformations at $t=5$ using 50 unroll steps with $\Delta t=0.02$. 
Top row: last observed input ($t=4$). Second row: forecasted deformation. Third row: ground truth at $t=5$.}
\label{fig:visual_forecast}
\end{figure}

Figure~\ref{fig:visual_forecast} presents a visual comparison between forecasted deformations and the ground truth at $t=5$ using $50$ unroll steps with $\Delta t = 0.02$. 
Each column shows one sequence: the last observed input at $t=4$ (top row), the predicted deformation at $t=5$ (second row), and the ground truth $t=5$ (third row). The Schr\"odinger-inspired forecasting model successfully captures both the global morphology and the fine-scale irregularities of the evolving structures.

\begin{figure}[thb!]
\centering
\includegraphics[width=0.6\linewidth]{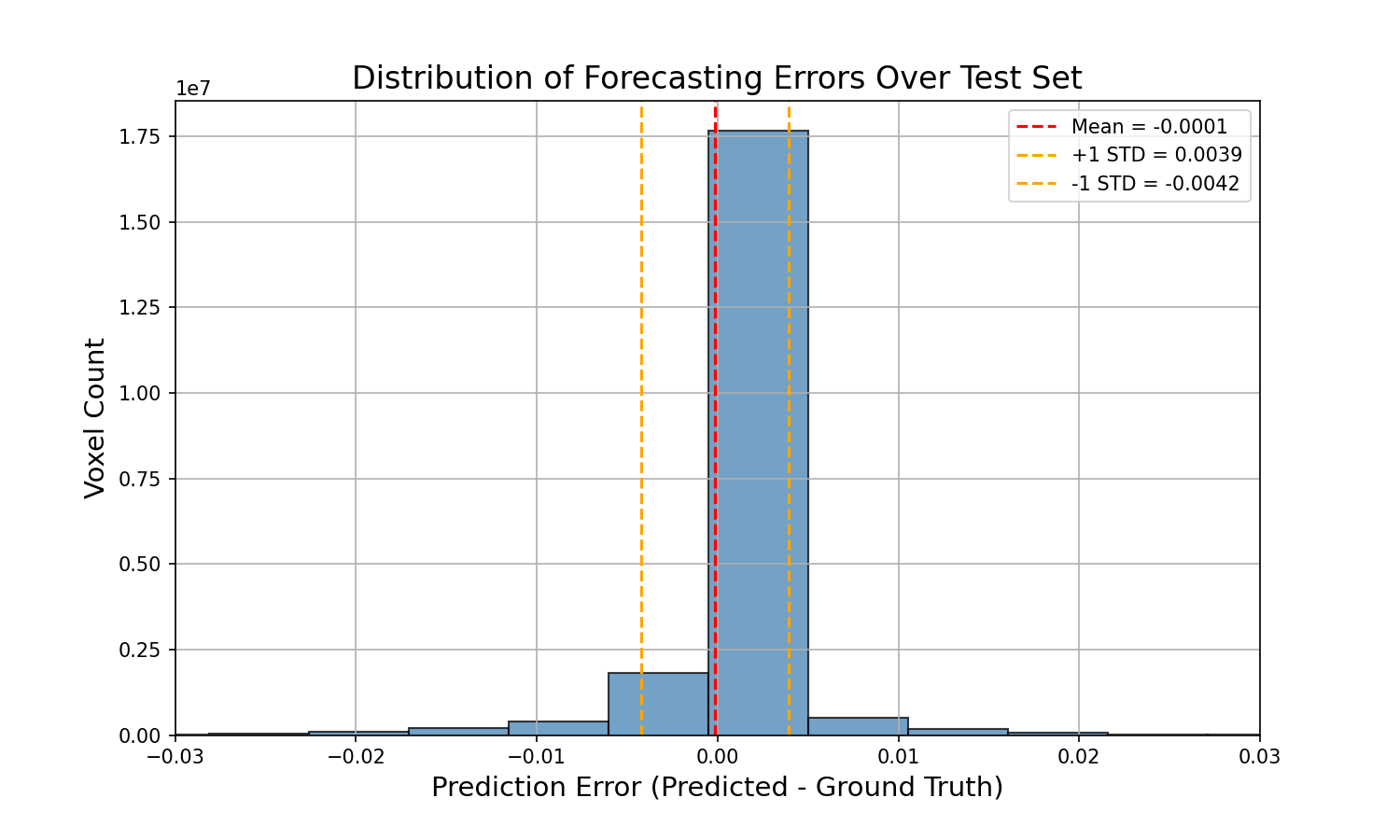}
\caption{Distribution of voxelwise forecasting errors across the test set. The error distribution is centered near zero with most errors confined to $\pm 0.004$, confirming unbiased and stable predictions.}
\label{fig:error_hist}
\end{figure}

Figure~\ref{fig:error_hist} shows the voxelwise distribution of forecasting errors (predicted minus ground truth intensity) across the test set. 
The distribution is sharply centered around zero with a mean error of $-1\!\times\!10^{-4}$, indicating the absence of systematic bias in forecasts. 
Most errors fall within one standard deviation ($\pm 0.004$), highlighting that deviations are small relative to the dynamic range of the volumes. 
The narrow, symmetric spread confirms that the proposed Schr\"odinger-inspired model not only achieves high global similarity scores but also preserves local accuracy at the voxel level, avoiding drift or structural distortions.

\begin{figure*}[thb!]
    \centering
    \includegraphics[width=0.32\linewidth]{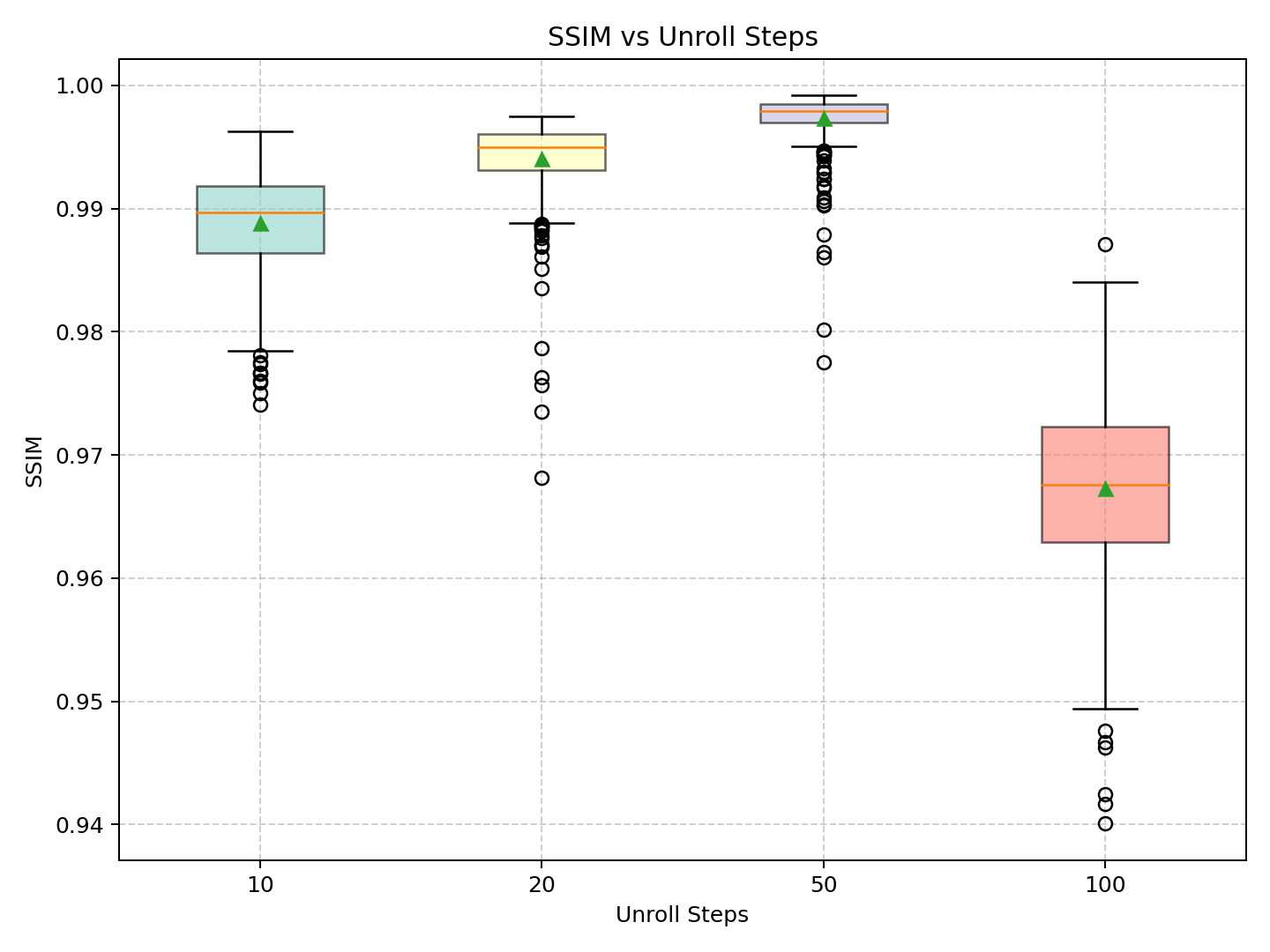}
    \includegraphics[width=0.32\linewidth]{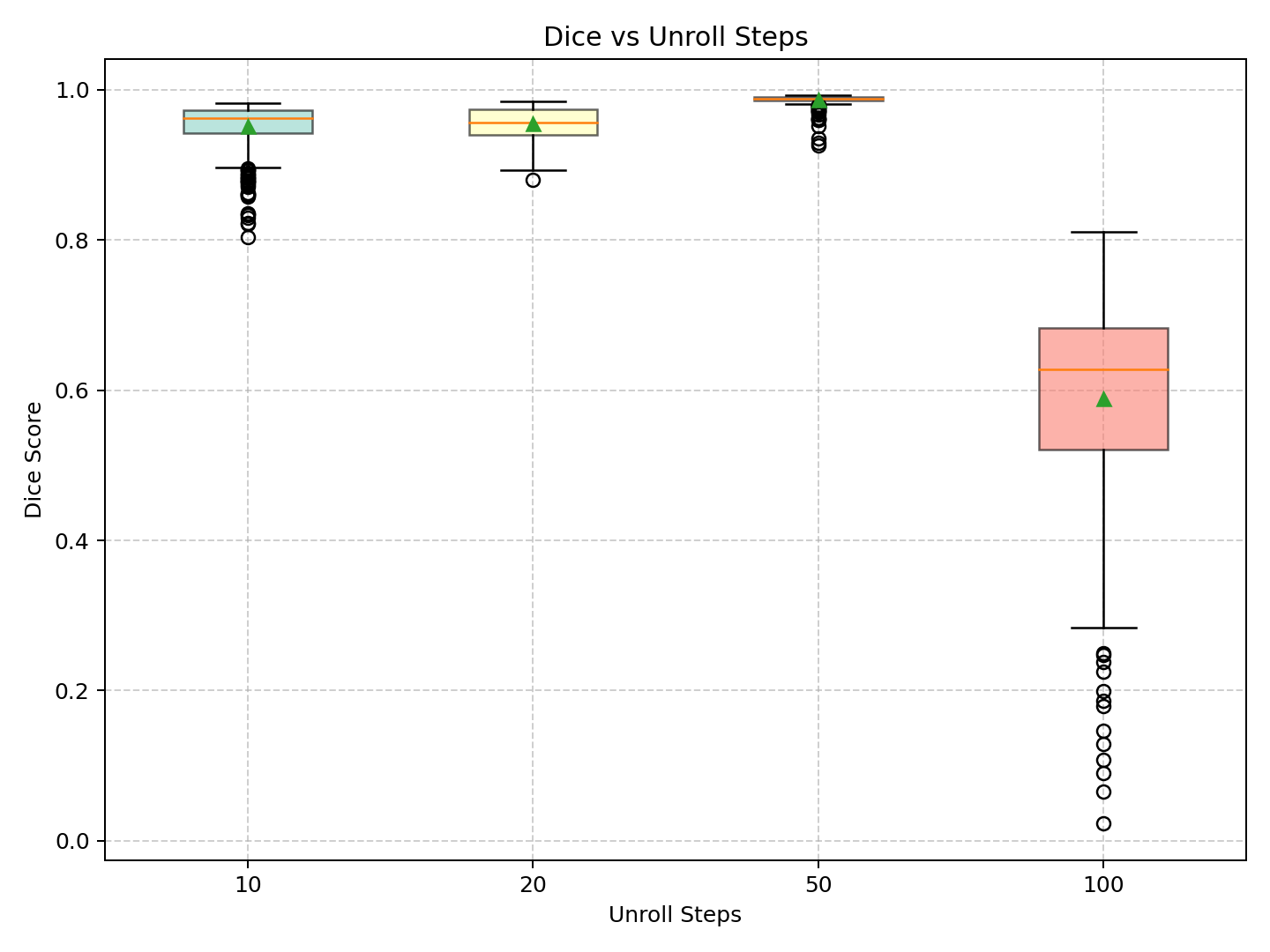}
    \includegraphics[width=0.32\linewidth]{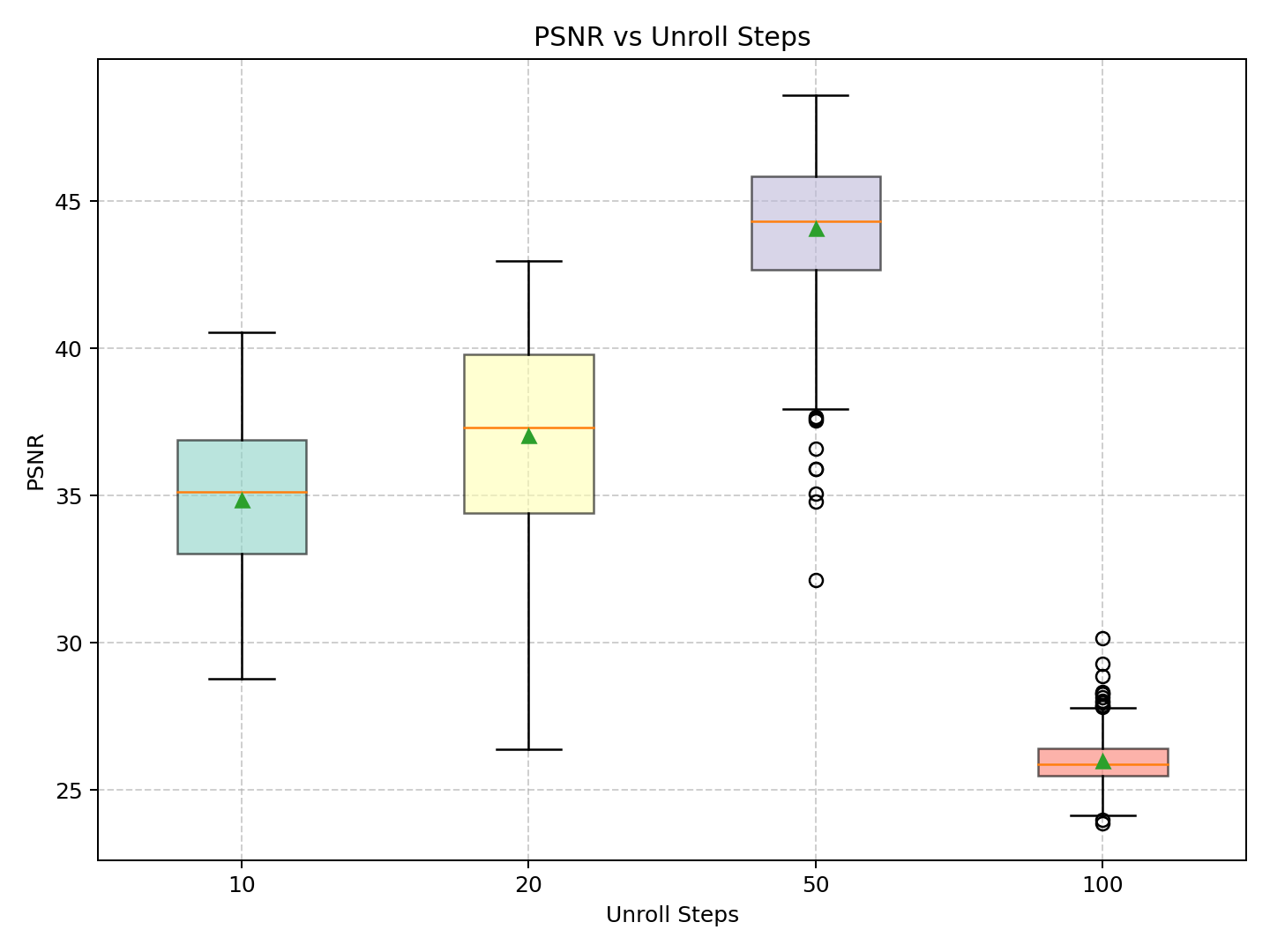}
    \caption{Boxplots of SSIM, Dice, and PSNR across different unroll steps (3D datasets).}
    \label{fig:boxplots_unroll}
\end{figure*}

As shown in Figure~\ref{fig:boxplots_unroll}, the number of unroll steps has a strong effect on forecasting accuracy. 
With $10$--$20$ steps, the model achieves reasonable performance but exhibits higher variance across test samples. 
At $50$ unrolls, the method reaches peak performance in all three metrics (SSIM, Dice, PSNR), providing the best balance of accuracy and stability. 
When extended to $100$ steps, however, forecasting accuracy declines, reflecting accumulated numerical drift during long temporal integration. 
These results confirm that $50$ unroll steps provide the most reliable forecasts, whereas both under- and over-unrolling reduce performance.

\subsection{Interpretability Through Schr\"odinger-Inspired Latent Fields}

\begin{figure}[thb!]
\centering
\includegraphics[width=0.8\linewidth]{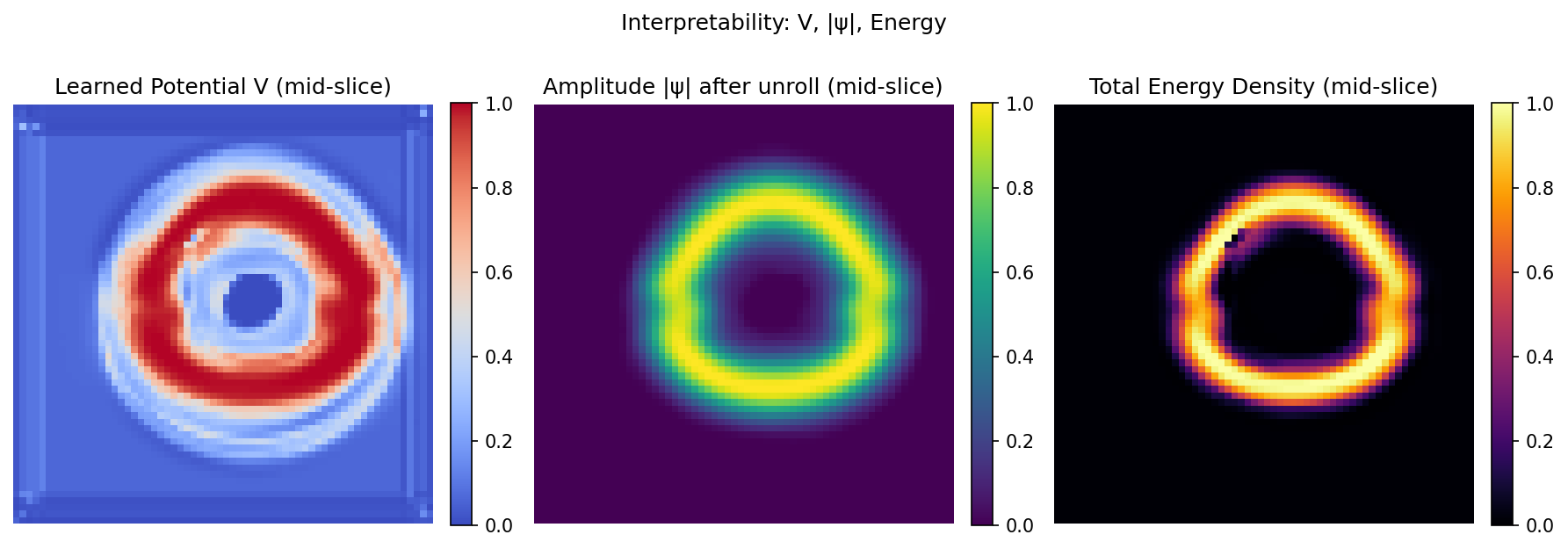}
\caption{Interpretability analysis of learned latent fields. 
Left: potential $V$ (mid-slice), indicating regions driving the deformation. 
Middle: amplitude $|\psi|$ after unrolling, capturing the evolving morphology. 
Right: total energy density, highlighting regions of concentrated dynamical activity.}
\label{fig:interpret_fields}
\end{figure}

A key advantage of the proposed Schr\"odinger-inspired forecasting model is that its latent representation is physically structured and interpretable. 
Figure~\ref{fig:interpret_fields} visualizes the learned potential $V$, the amplitude $|\psi|$ after unrolled evolution, and the total energy density for a representative test case. 
The potential field $V$ highlights regions that act as deformation drivers, with high-intensity zones indicating where growth or motion forces are most concentrated. 
The amplitude field $|\psi|$ closely aligns with the structural morphology of the predicted deformation, effectively capturing the evolving boundaries of the object. 
The derived energy density, computed as $|H\psi|^2$, provides a complementary measure that localizes regions of high dynamical activity. 
Together, these views reveal that the model not only outputs predictions but also provides meaningful intermediate fields that can be inspected to understand how and where deformations are being driven.

\begin{figure}[thb!]
\centering
\includegraphics[width=0.8\linewidth]{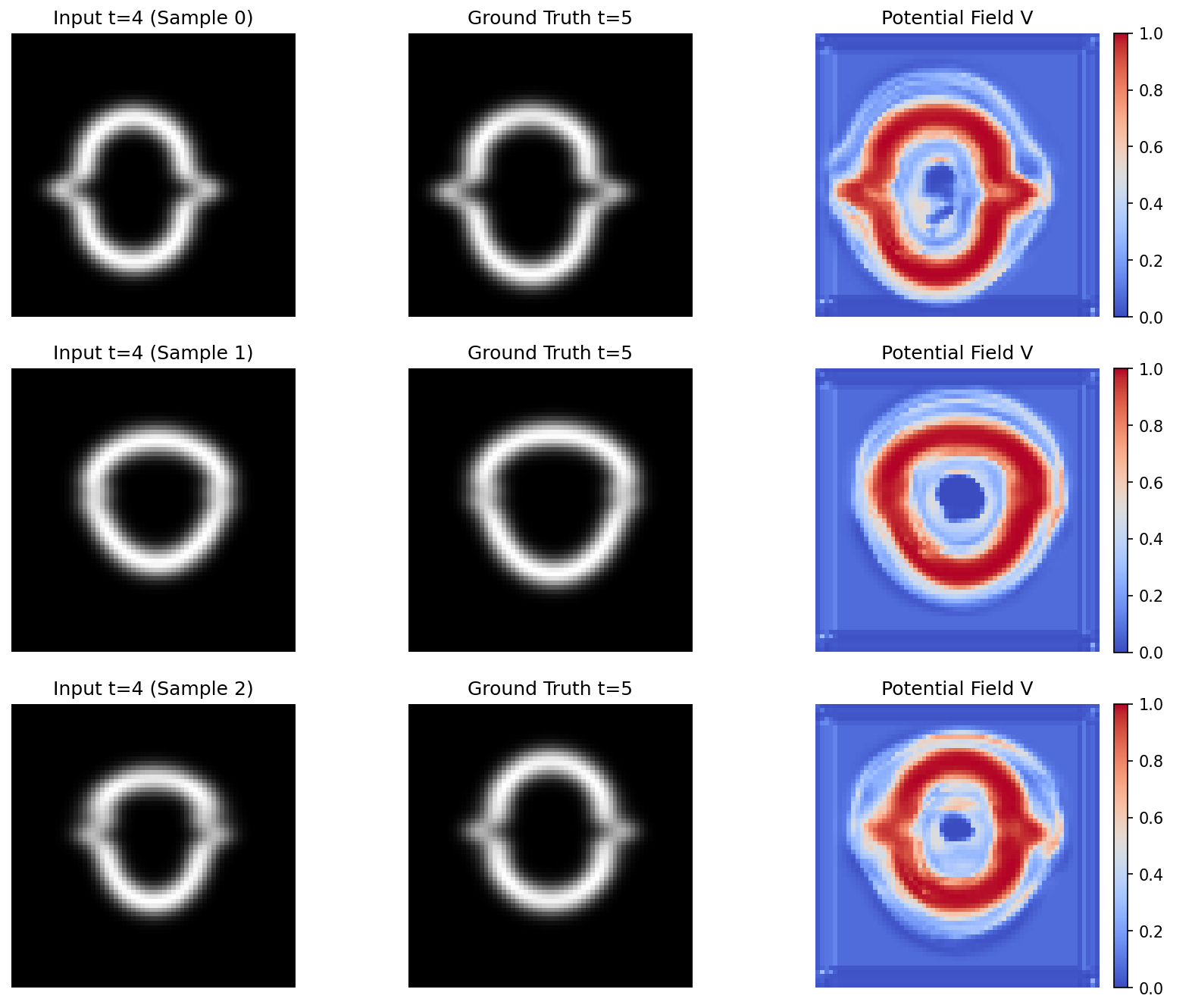}
\caption{Examples of interpretability across sequences. 
Each row shows the input at $t=4$ (left), ground truth at $t=5$ (middle), and the learned potential field $V$ (right). 
The potential consistently highlights boundary regions driving the deformation, offering interpretable insights into the model’s internal dynamics.}
\label{fig:interpret_samples}
\end{figure}

We further examine how the learned potential adapts across different input sequences. 
Figure~\ref{fig:interpret_samples} shows three representative cases, comparing the last observed input at $t=4$, the ground truth at $t=5$, and the corresponding potential field. Despite differences in shape and deformation dynamics, the learned potential consistently localizes near the evolving boundaries, acting as a spatial regulator for the forecast. This demonstrates that the potential field serves as an interpretable driver of the predicted motion, bridging the gap between black-box learning and physically guided modeling.

\subsection{Comprehensive Evaluation of Volumetric Accuracy, Efficiency, and Structural Fidelity}

Tables~\ref{tab:volume_compute} and \ref{tab:spectral_surface} provide a comprehensive evaluation of forecasting quality across different unroll depths. 
The first table reports global volumetric accuracy and computational efficiency. 
At $50$ unrolls, the predicted tumor volume closely matches the ground truth (absolute error below $1\%$) and the center-of-mass (CoM) deviation is only $0.05$ voxels, demonstrating excellent preservation of global shape and mass balance. 
This level of accuracy is achieved with modest computational overhead: latency increases only slightly compared to $10$--$20$ unrolls, while GPU memory remains constant. 
In contrast, $100$ unrolls accumulate numerical drift, producing large underestimation of volume (over $55\%$ error) and degraded stability.

The second table captures finer-grained quality measures: spectral error decomposition and surface-based accuracy. 
The spectral analysis separates forecasting errors into high-frequency (fine structures and boundaries) and low-frequency (global smooth trends) components. 
Here, $50$ unrolls maximize the retention of high-frequency content ($39.4\%$ of error attributed to fine details), indicating sharper preservation of structural boundaries compared to both shorter (10--20) and longer (100) unrolls, which either oversmooth or collapse into low-frequency drift. 

Surface-based metrics provide complementary geometric validation. 
The Hausdorff distance at 95th percentile (HD95) and the average symmetric surface distance (ASSD) measure spatial deviations between predicted and ground-truth boundaries, while Surface Dice@1\,vox quantifies overlap when surfaces are dilated by one voxel tolerance. 
These metrics are particularly relevant in medical imaging, where accurate boundary prediction is critical for treatment planning and anatomical analysis. 
Results show that $50$ unrolls dramatically reduce HD95 and ASSD errors and achieve near-perfect surface overlap (Surface Dice $=0.9999$). 
At 100 unrolls, however, both distance errors increase substantially and overlap drops below $0.82$, confirming loss of geometric fidelity.

Taken together, these results highlight the importance of moderate unrolling: $50$ steps provide the optimal balance between global accuracy, structural sharpness, boundary alignment, and computational efficiency. 
This operating point enables the Schr\"odinger-inspired model to deliver both volumetric and surface-consistent forecasts, which is essential for downstream applications in 4D medical imaging and scientific simulations.

\begin{table}[t]
\centering
\caption{Spectral error decomposition and surface accuracy across unroll steps. Best values are in bold.}
\label{tab:spectral_surface}
\resizebox{\columnwidth}{!}{
\begin{tabular}{c|cc|ccc}
\toprule
\multirow{2}{*}{Unroll} 
& \multicolumn{2}{c|}{\textbf{Spectral Error}} 
& \multicolumn{3}{c}{\textbf{Surface Accuracy}} \\
\cmidrule(lr){2-3} \cmidrule(lr){4-6}
& High-Freq Err & Low-Freq Err 
& HD95 (vox) & ASSD (vox) & Surface Dice@1\,vox \\
\midrule
10  & 0.231 & 0.769 & 1.000  & 0.189 & 0.9988 \\
20  & 0.316 & 0.684 & 1.000  & 0.181 & 0.9996 \\
50  & \textbf{0.394} & \textbf{0.606} & \textbf{0.347} & \textbf{0.052} & \textbf{0.9999} \\
100 & 0.041 & 0.959 & 1.864 & 1.111 & 0.8196 \\
\bottomrule
\end{tabular}}
\end{table}

\begin{table}[t]
\centering
\caption{Volume accuracy and compute profile across unroll steps. Best values are in bold.}
\label{tab:volume_compute}
\resizebox{\columnwidth}{!}{
\begin{tabular}{c|cccc|ccc}
\toprule
\multirow{2}{*}{Unroll} 
& \multicolumn{4}{c|}{\textbf{Volume Accuracy}} 
& \multicolumn{3}{c}{\textbf{Compute Profile}} \\
\cmidrule(lr){2-5} \cmidrule(lr){6-8}
& Pred. Vol & GT Vol & Abs. Vol Err (\%) & CoM Err 
& Latency (ms) & Throughput (vol/s) & GPU (MB) \\
\midrule
10  & 12289.3 & 13102.8 & 7.94  & 0.15 & 62.6 & \textbf{15.97} & 252.8 \\
20  & 13935.7 & 12858.6 & 8.32  & 0.10 & 63.0 & 15.88 & 252.8 \\
50  & \textbf{12766.7} & 12870.0 & \textbf{0.92} & \textbf{0.05} & 69.1 & 14.47 & 252.8 \\
100 & 5733.1  & 12879.7 & 55.85 & 0.60 & 85.9 & 11.64 & 252.8 \\
\bottomrule
\end{tabular}}
\end{table}

\begin{figure}[h!]
    \centering
    \includegraphics[width=0.7\textwidth]{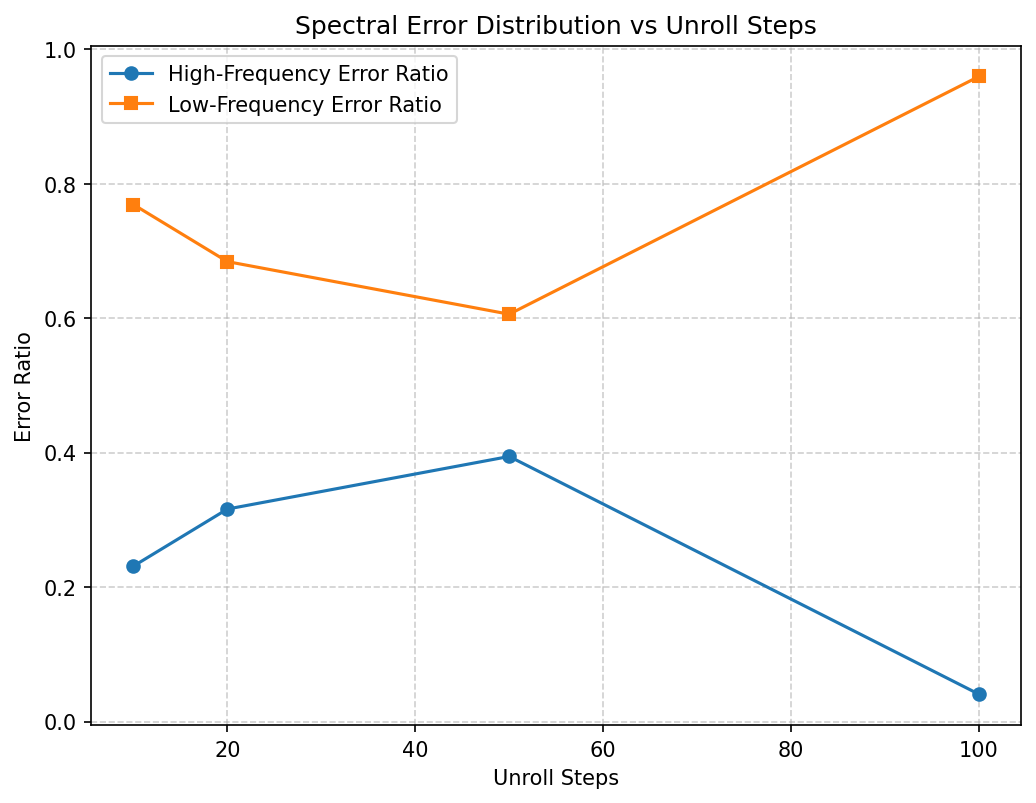}
    \caption{Spectral decomposition of forecasting errors across unroll steps. 
    Low-frequency errors (orange) correspond to global shape drift, while high-frequency errors (blue) capture boundary irregularities. 
    Results show that moderate unrolls (e.g., 50 steps) achieve the best trade-off, minimizing both components simultaneously.}
    \label{fig:spectral_error}
\end{figure}

Figure~\ref{fig:spectral_error} shows the spectral decomposition of forecasting errors across different unroll steps. The errors are separated into low-frequency components (orange) capturing smooth, global shape discrepancies, and high-frequency components (blue) capturing fine boundary irregularities. At moderate unroll steps (e.g., 50), the balance between low- and high-frequency errors is most favorable, with reduced low-frequency drift and controlled high-frequency deviations. In contrast, very short horizons (10 steps) yield elevated low-frequency errors due to insufficient temporal modeling, while excessively long horizons (100 steps) increase low-frequency drift even though high-frequency error drops sharply. This analysis highlights that the proposed method maintains a strong trade-off between global shape preservation and boundary fidelity, particularly at moderate unroll depths.

\subsection{Generalization of Schr\"odinger-Inspired Forecasting to 2D Data}

\begin{figure}[h!]
    \centering
    \includegraphics[width=0.95\linewidth]{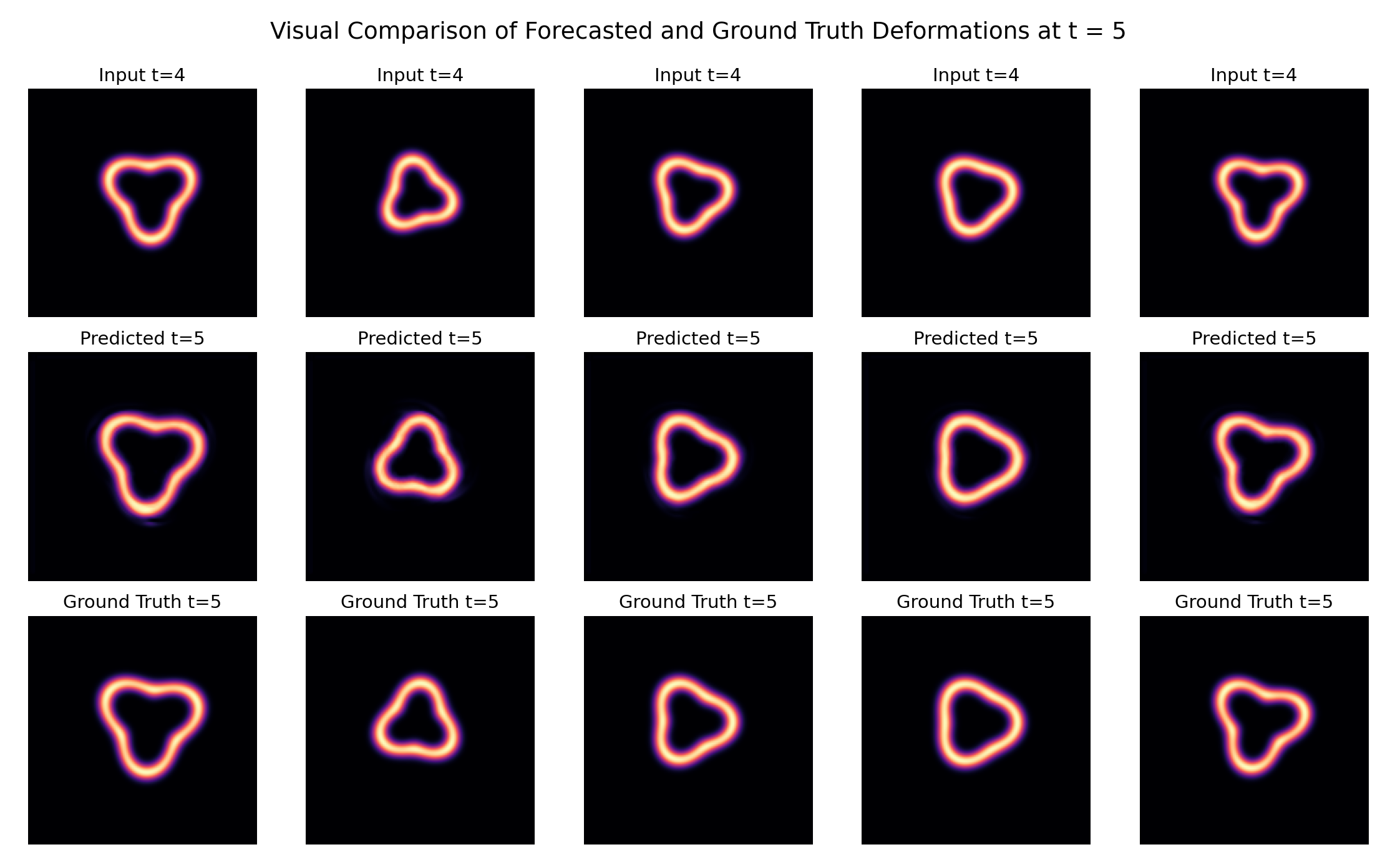}
    \caption{Visual comparison of forecasted and ground-truth 2D deformations at $t=5$. 
    The model successfully captures both morphological evolution and rotational changes, demonstrating its generalization beyond 4D volumetric data.}
    \label{fig:forecast2d}
\end{figure}

Although our primary focus is on 4D volumetric forecasting, we also evaluated the proposed Schr\"odinger-inspired methodology on 2D synthetic datasets to verify its generalizability. 
The results demonstrate that the model effectively captures deformation dynamics in two dimensions, including rotational motion, as shown in Figure~\ref{fig:forecast2d}. 
The ability to maintain shape fidelity while accurately predicting rotational deformations highlights the robustness of the proposed representation even outside the full 4D setting.

\noindent
Table~\ref{tab:metrics2d} summarizes SSIM, PSNR, MSE, and Dice scores for different unroll depths on the 2D dataset. 
Consistent with the 4D experiments, the 2D results show that performance improves with deeper unrolling up to 50 steps, where the model achieves the highest PSNR (48.27 dB), lowest MSE ($2.0\times10^{-5}$), and best Dice score (0.9915). 
Excessive unrolling (100 steps) leads to error accumulation, reducing both fidelity (PSNR = 40.73) and overlap accuracy (Dice = 0.9855). 
These results confirm that the proposed framework maintains robustness across dimensionalities while retaining similar stability and interpretability properties.

\begin{table}[h!]
\centering
\caption{2D Forecasting Quality Metrics Across Unroll Steps}
\label{tab:metrics2d}
\begin{tabular}{c|c|c|c|c}
\hline
\textbf{Unroll Steps} & \textbf{SSIM} & \textbf{PSNR (dB)} & \textbf{MSE} & \textbf{Dice} \\
\hline
10  & 0.9970 & 42.01 & 0.000070 & 0.9867 \\
20  & 0.9976 & 43.69 & 0.000049 & 0.9886 \\
50  & 0.9987 & 48.27 & 0.000020 & 0.9915 \\
100 & 0.9973 & 40.73 & 0.000093 & 0.9855 \\
\hline
\end{tabular}
\end{table}

\section{Conclusion}
In this work, we presented a Schr\"odinger-inspired 4D deformation forecasting model that embeds the dynamics of the time-dependent Schr\"odinger equation within a deep learning framework. By disentangling amplitude, phase, and potential fields, our approach introduces interpretable latent variables that govern deformation dynamics in a structured manner. The incorporation of an approximate Crank--Nicolson evolution ensures stable temporal propagation, mitigating drift and error accumulation over long horizons while remaining computationally tractable. Through extensive experiments on synthetic 3D and 4D datasets, we demonstrated that the proposed method achieves high fidelity in both reconstruction metrics (SSIM, PSNR, MSE) and segmentation accuracy (Dice score), with robustness to threshold selection and forecasting depth. The interpretability analyses further highlight the role of the learned potential field in guiding deformations, offering a physically meaningful perspective that is often absent in conventional black-box models. This approach has potential applications in medical imaging (e.g., tumor growth modeling, organ motion tracking), materials science (structural evolution), and geophysics (fluid or seismic deformation forecasting). Future work will extend this framework to real-world clinical datasets, incorporate stronger physical regularization, and explore hybrid integration with biomechanical models to further enhance robustness and realism.

\bibliographystyle{abbrv}
\bibliography{references}
\end{document}